\documentclass[runningheads]{llncs}
\usepackage[T1]{fontenc}

\usepackage{graphicx}
\usepackage{multirow}

\usepackage{hyperref}
\usepackage{booktabs}
\usepackage{amsmath} 
\usepackage{comment}

\usepackage[export]{adjustbox}
\usepackage{tcolorbox}
\usepackage{hyperref}
\usepackage{xcolor} 
\usepackage{color}

\begin{document}
%
\title{LML: A Novel Lexicon for the Moral Foundation of Liberty}%
\titlerunning{LML: Liberty Moral Lexicon}

\author{Oscar Araque\inst{1}\orcidID{0000-0003-3224-0001} \and Lorenzo Gatti\inst{2}\orcidID{0000-0003-2422-5055} \and Sergio Consoli\inst{3}\orcidID{0000-0001-7357-5858} \and Kyriaki Kalimeri\inst{4}\orcidID{0000-0001-8068-5916} }
\institute{Universidad Polit\'ecnica de Madrid, Spain \email{o.araque@upm.es} \and University of Twente, Enschede, The Netherlands \email{l.gatti@utwente.nl} \and European Commission, Joint Research Centre (DG JRC), Ispra, Italy \email{sergio.consoli@ec.europa.eu} \and ISI Foundation, Turin, Italy \email{kyriaki.kalimeri@isi.it} }

\authorrunning{O. Araque et al.}

\maketitle 
\begin{abstract}
The moral value of liberty is a central concept in our inference system when it comes to taking a stance towards controversial social issues such as vaccine hesitancy, climate change, or the right to abortion. 
Here, we propose a novel Liberty lexicon evaluated on more than 3,000 manually annotated data both in in- and out-of-domain scenarios.
As a result of this evaluation, we produce a combined lexicon, referred to as the Liberty Moral Lexicon (LML), that constitutes the main outcome of this work.
LML incorporates information from
an ensemble of lexicons that have been generated
using word embedding similarity (WE) and compositional semantics (CS).
Our key contributions include enriching the liberty annotations, developing a robust liberty lexicon for broader application, and revealing the complexity of expressions related to liberty 
across different platforms.
Through the evaluation, we show that the difficulty of the task calls for designing approaches that combine knowledge, in an effort of improving the representations of learning systems.
\keywords{Natural language, moral foundations, liberty, oppression.}
\end{abstract}




\section{Introduction}
Moral values are fundamental to our decision-making process, especially regarding controversial social issues. When taking a stance, for instance, on global warming or vaccine adherence, we consult - consciously or unconsciously - our moral system of values. 
The Moral Foundations Theory (MFT) was created precisely to explain morality across cultures \cite{Haidt2004}, proposing five foundations, namely \textit{care, fairness, loyalty, authority} and \textit{sanctity}. In a much later revision, the theory was enhanced with a new sixth dimension: \textit{liberty}~\cite{haidt2012righteous}.
The ``Liberty/Oppression'' foundation is about people's reactance and resentment towards those who dominate them and restrict their liberty. 
Moral notions captured by the Liberty foundation include freedom of choice and individual responsibility of actions, which repeatedly emerged as fundamental decision-making drivers of crucial prosocial behaviors such as vaccine adherence~\cite{Amin_2017,beiro2023moral,ZHANG2023107479}, and cooperation during crisis~\cite{mejova2023authority}.
%
Recent works focus on the automatic detection of moral values in text, employing annotated lexicons, either for unsupervised detection or as features in a learning system \cite{Mooijman2018,rezapour2021incorporating,kennedy2021moral,preniqi2021modelling,mejova2023authority,ZHANG2023107479}. 
Given that the \textit{liberty} foundation was added to the MFT theory subsequently, there were initially no linguistic resources for it. 
Preliminary approaches to liberty assessment were based on purely data-driven lexical characterization~\cite{araque2021liberty,araque2022libertymfd}, lacking, however, a solid evaluation against a benchmarked ground-truth.
This is precisely the gap we are addressing in this study.

We gathered data from various platforms to cover multiple aspects of the expression of the liberty foundation. 
In particular, we included (i) the \textit{Wikipedia}\footnote{\href{https://www.wikipedia.org}{https://www.wikipedia.org}} and \textit{Conservapedia}\footnote{\href{https://www.conservapedia.com/Main_Page}{https://www.conservapedia.com}} projects, encyclopedia projects of general content but diverse viewpoints, (ii) the \textit{r/Libertarian} and \textit{r/Conservative} communities on \textit{Reddit.com}, forums of political discussion of general interest, (iii) the Black Lives Matter (BLM) and Election (Elect.) datasets from the \textit{MFTC Twitter Corpus}~\cite{hoover2020moral}, a collection of tweets discussing racial discrimination and the US Presidential Elections of 2016, respectively, as well as (iv) posts and comments from \textit{META's Pages} regarding the vaccination debate (Vaccine).
The first two scenarios (Wikipedia vs Conservapedia and Libertarian vs Conservative on Reddit) act as ``natural experiments'' expressing the viewpoints of communities with diverse opinions and stances on the liberty foundation described by the MFT framework. 
To ensure a robust ground-truth, we obtained manual annotations of the expression of the moral foundation of liberty in the BLM, Elect., and META's vaccination-related posts and comments.

We generated two lexicons per dataset, employing two complementary approaches; the word embedding similarity (WE)~\cite{turney2003measuring} and the compositional semantics (CS)~\cite{liang2013learning}.
The first automatically extracts a set of seed words using frequency shifts, comparing new words' embeddings to seed words' embeddings to determine their alignment with the foundation's principles. 
The second method assumes that each word expresses the side of the foundation more frequently present in the documents where the word appears.
We explore a lexicon aggregation approach based on overlapping terms that combines the benefits of the two methods, resulting in the \textit{Liberty Moral Lexicon (LML)}.
We also propose a combined representation approach which takes into consideration the individual lexical resources while accounting for overfitting issues.
Finally, we evaluate the lexicons obtained per dataset, both in cross-domain experimental setups and out-of-domain ones.

We contribute to the state of the art in moral understanding by augmenting the benchmark dataset of the Moral Foundations Theory Corpus (MFTC) with valuable manual annotations on the liberty moral foundation and rendering them available to the scientific community. Our novel Liberty Moral Lexicon 
combines a variety of data sources covering a wide range of topics, resulting into a refined and versatile liberty lexicon capable of effectively generalizing over previously unseen domains\footnote{The LML dictionary is available at the following link: \url{https://github.com/oaraque/moral-foundations/tree/master/liberty/3rd_version}}. 
Moreover, 
LML enables researchers to bridge qualitative and quantitative methods, enhancing tools like TOMEA~\cite{liscio2023} in uncovering the contextual variability of moral language.
These contributions collectively enhance our understanding of moral analysis and pave the way for more accurate and comprehensive evaluations.


\section{Related Works}

The Moral Foundations Dictionary (MFD)~\cite{graham2009liberals} is a collection of lemmas and associated moral traits, assembled by experts and typically used together with the Linguistic Inquiry and Word Count (LIWC) software~\cite{Tausczik2010} to estimate moral traits and investigate differences in moral concerns between different cultural groups.
Garten et al. \cite{Garten2018} proposed the Distributed Dictionary Representations (DDR) method based on psychological dictionaries and semantic similarity to quantify the presence of moral sentiment around a given topic. 
Later on, the authors extended the method, incorporating demographic embeddings into the language representations~\cite{Garten2019}.

In an attempt to address several of the limitations of the MFD, \cite{araque2020moralstrength} proposed a data-driven generated lexicon, the \textit{MoralStrength}, which expanded the original MFD employing the WordNet synsets and crowdsourced annotations. Different from the MFD, where each foundation is considered a bipolar of ``virtue'' and ``vice'', MoralStrength treats each foundation as a continuum, assigning a numeric value of moral valence to each lemma that indicates the weight with which the lemma is expressing the specific value. 
\cite{hopp2021extended} developed the \textit{extended Moral Foundations Dictionary} (eMFD), a lexicon which expands the MFD based on crowdsourced annotations. Each lemma in eMFD is assigned a continuously weighted vector that expresses the probability that the lemma belongs to any of the five moral foundations. 

Notably, none of the above lexicons though included the liberty moral foundation. 
A first attempt to derive a lexicon from assessing the presence of liberty in the text was presented by \cite{araque2021liberty}. They considered pairs of Wikipedia Pages and their Conservapedia counterparts as natural expressions of the liberty-oppression divide. They created a series of word embeddings which were then compared through cosine similarity to a set of seed words defined by experts to generate a lexicon. 
Their design comes with the obvious conceptual limitations of considering the Wikipedia project as expressing a strongly libertarian position and initiating the embeddings with a list of manually selected seed words from expert annotators. 
More recently, ~\cite{araque2022libertymfd}
proposed a liberty lexicon generation approach based 
aligning documents from online news sources with different worldviews. 
The LibertyMFD was later employed in~\cite{araque2023beyond} to fine-tune the approach proposed by Consoli et al.~\cite{Consoli2022a} for analysing how the Spanish news cover the female (un)employment topic in terms of sentiment and moral values, as well as how this sentiment evolves over time.

Although pioneering, their approach suffered from the lack of a solid ground-truth on which to evaluate the generated lexicons. Due to the lack of a ground-truth dataset, the lexicon evaluation was based on the assumption that news from different political orientations would express opposite notions with respect to the \textit{liberty} moral foundation.
Instead, in LML, 
we evaluate the lexicons against solid manual annotations for the \textit{liberty} foundation of the benchmark Moral Foundations Twitter Corpus datasets (BLM and Elect.) which we render available to the scientific community. 

\section{Data Collection}

\paragraph{Moral Foundations Twitter Corpus (MFTC).} The MFTC is a corpus consisting of seven independent datasets (35k tweets in total), manually annotated for the original five moral foundations~\cite{hoover2020moral}, but not the \textit{liberty} foundation. The Black Lives Matter Twitter Corpus (BLM) and the Elections Corpus (Elect.) are the two largest datasets in this collection, and we manually annotated them\footnote{We did not proceed to the annotation of the entire MFTC corpus due to funding limitations.} as per the moral foundation of Liberty, relying on a popular tool for crowdsourcing and human validation, i.e., Amazon Mechanical Turk (AMT hereafter) provided by the Amazon SageMaker Ground Truth service. 
The Black Lives Matter Twitter Corpus focuses on tweets specifically regarding the Black Lives Matter movement, and it contains 4,352 tweets. The Election corpus relates to the US 2016 Presidential election, and consists of 4,370 tweets. 
To ensure coherence, we followed the same procedure and annotation scheme of Hoover et al.~\cite{hoover2017moral} used for the MFTC annotation. 
Moreover, inspired by the annotation approach of the MoralStrength lexicon~\cite{araque2020moralstrength}, we added the notion of ``strength'', which indicates the degree to which each lemma expresses the \textit{liberty} moral foundation in addition to its presence and polarity.
We assigned each tweet to nine independent annotators and asked them to rate the extent to which each tweet expressed a ``Liberal'' or ``Oppressive'' moral value on a scale from 1 to 9. 
The score magnitude represents the intensity of the Liberty/Oppression expressed in a tweet, as perceived by the annotator: 
a score close to 9 indicates that the sentence expresses a highly oppressive connotation, while a score value close to 1 is associated with a very libertarian connotation. Should the sentence not be associated with neither an oppressive nor a libertarian connotation, then the annotator could assign a neutral score. The intercoder agreement score provided by AMT is 92\%\footnote{Annotations will be available upon acceptance.}.

\begin{table*}
 \centering
 \resizebox{\textwidth}{!}{
 \begin{tabular}{l|rrr|rrr}
 \toprule
 \multirow{2}{*}{\textbf{Dataset}} & \multicolumn{3}{c|}{\textbf{Original}} & \multicolumn{3}{c}{\textbf{Balanced}} \\
 & \textbf{Label} & \textbf{Instances (\%)} & \textbf{Total instances} & \textbf{Label} & \textbf{Instances (\%)} & \textbf{Total instances} \\
 \midrule
 \multirow{3}{*}{\shortstack[l]{Black Lives \\Matter (BLM)}} & Liberty & 54\% & \multirow{3}{*}{4,340} & \multirow{3}{*}{\shortstack[r]{Liberty/Oppression\\Neutral}} & \multirow{3}{*}{\shortstack[r]{50\% \\50\%}} & \multirow{3}{*}{1,600}\\
 & Neutral & 18\% & \\
 & Oppression & 28\% & \\
 \midrule
 \multirow{3}{*}{Election} & Liberty & 56\% & \multirow{3}{*}{4,366} & \multirow{3}{*}{\shortstack[r]{Liberty/Oppression\\Neutral}} & \multirow{3}{*}{\shortstack[r]{50\% \\50\%}} & \multirow{3}{*}{1,532} \\
 & Neutral & 18\% & \\
 & Oppression & 26\% & \\
 \midrule
 \multirow{2}{*}{Reddit} & Libertarian & 51\% & \multirow{2}{*}{100,000} & Libertarian & 51\% & \multirow{2}{*}{100,000} \\
 & Conservative & 49\% & & Conservative & 49\% \\
 \midrule
 \multirow{2}{*}{\shortstack[l]{Wikipedia+\\Conservapedia}} & Libertarian & 50\% & \multirow{2}{*}{57,078} & Libertarian & 50\% & \multirow{2}{*}{57,078} \\
 & Conservative & 50\% & & Conservative & 50\% \\
 \midrule
 \multirow{2}{*}{Vaccination} & Liberty/Oppression & 89\% & \multirow{2}{*}{1,576} & Liberty/Oppression & 50\% & \multirow{2}{*}{356}\\
 & Neutral & 11\% & & Neutral & 50\% \\
 \bottomrule
 \end{tabular}
 }
 \caption{Overview of the datasets used in this work. Generation of the lexicons is performed on the Original version of the datasets while training of the regression models is performed on the Balanced version of the datasets.}
 \label{tab:datasets}
\end{table*}

\paragraph{Reddit.}
Reddit is increasingly becoming a reliable data source in computational studies~\cite{proferes2021reddit}.
Aiming to profile the language of libertarian and conservative users, we extracted textual content from the \texttt{r/Libertarian} and \texttt{r/Conservative} communities, which are self-proclaimed networks of libertarian and conservative ideas, respectively.
Initially, we considered posts and comments published between August 2008 and April 2021, obtaining overall 1,127,005 documents.
From these, we have filtered empty and other unusable content, and undersampled the rest to obtain a final amount of 100,000 instances.

\paragraph{Wikipedia+Conservapedia.}
We use the dataset described in~\cite{araque2021liberty}, based on page alignment between Wikipedia and Conservapedia according to their title (henceforth the WikiCon dataset).
More than 37,000 articles between Wikipedia and Conservapedia have been aligned, of which approximately 28,000 pages had identical titles, and the remaining were aligned based on redirect pages.
The entire corpus contains 106 million tokens and 558,000 unique words.
The dataset has been filtered using page categories related to politics, while a length ratio filter has been applied between Wikipedia and Conservapedia documents to improve dataset quality.
This ratio compares the number of words in a Wikipedia document to the number of terms in the corresponding Conservapedia document, and excludes the pairs with ratio higher than 10, resulting in 57,078 documents split equally between 28,539 Conservapedia and Wikipedia sources.


\paragraph{Vaccination.} 
Finally, we use a dataset on vaccinations, which comprises anonymous posts and comments from about 200 Facebook Pages, collected through the Facebook API from January 2012 to June 2019~\cite{beiro2023moral}.
The total number of comments and posts from both sides of the vaccine debate amounts to 607,105.
The creators of the dataset randomly selected approximately 1,500 comments and manually annotated the presence of the liberty moral foundation in the snippet, indicating also the polarity of the foundation as ``virtue'' (liberty) or ``vice'' (oppression).
A summary of the datasets used in this work is in Table~\ref{tab:datasets}.

\section{Methods \& Evaluation}

\subsection{Data Preprocessing}
A basic preprocessing was performed for all datasets, consisting of the following steps: stop words removal, token normalization, punctuation filtering, and removal of short words (i.e., terms with less than three letters).
Additionally, since the original datasets have slightly different annotations schemes as seen in Table~\ref{tab:datasets}, we aligned them, creating a binarised and balanced version of each dataset aggregating the labels accordingly. 
The binarisation process was performed by aggregating the Liberty and Oppression labels, thus creating a dataset where the annotation is either ``expresses liberty/oppression'' or ``doesn't express this moral foundation'', then balancing the classes by randomly undersampling the most populated class to match the population of the smaller class. 


\subsection{Lexicon Generation}
\paragraph{Word Embedding Similarity.}
\label{subsec:word-embedding-similarity}
Based on the approach proposed in \cite{turney2003measuring}, our first strategy for generating lexicons relies on word embedding similarity between the vectors of the positive and negative instances of a dataset's documents.
Hence, the method relies on a set of seed words that accurately represent the domains we aim to differentiate.
However, arbitrary selection of seed words can bias the output, since variations in the seed word list lead to differences in the final lexicon. 
To overcome this issue, we obtained the set of seed words in a data-driven way by estimating the frequency shifts~\cite{gallagher2021generalized} of the lemmas between the positive and negative documents, as done in~\cite{araque2022libertymfd}.
This approach helps us to avoid the limitations of arbitrarily selecting the seed words.
Thus, we consider the relative frequency of a word $w$ in a set of documents $D$:
\begin{equation}
 p_w^{(D)} = \frac{f_w^{(D)}}{\sum_{w' \in W^{(D)}} f_{w'}^{(D)}}
\end{equation}

\noindent where $w' \in W^{(d)}$ are the words in vocabulary set $W^{(D)}$ except for $w$.
We compute the frequency shift with relation to the relative frequency per word $w$ between two different sets of documents as:
\begin{equation}
\delta p_w = p_w^{(2)} - p_w^{(1)} 
\end{equation}

The seed word lists are generated based on prominent differences in word frequency shifts.
We apply a minimum frequency threshold of 100 to filter out less common lemmas.
We then use the word2vec algorithm~\cite{le2014distributed} to compute the vector for each word, using the standard parameter setting and a vector dimension of 300.
The lexicon is generated by estimating the cosine similarity between the word vectors obtained using the emerging seed words.
To compute the moral polarity of a word $w_i$ from the documents, we use the sets of seed words for the ``oppressive'' orientation ($S_C$) and the ``liberty'' direction ($S_L$), and estimate the polarity based on the cosine similarity:
\begin{equation}
\sum_{w_j \in S_L} \text{sim}(w_i, w_j) - \sum_{w_k \in S_C} \text{sim}(w_i, w_k)
\end{equation}
where \textit{sim} represents the cosine similarity as estimated by the word embedding model.
The obtained polarity is positive if $w_i$ is related to the positive seed words and a negative value if the word is more related towards the negative seed words.
We refer this model to as the \textit{WE model} and generate one lexicon for each dataset (except Vaccine, which is only used for testing).


\paragraph{Compositional Semantics.}
\label{subsec:compositional-semantics}

The second approach involves using the Compositional Semantics (CS) method~\cite{liang2013learning}, previously used to generate emotion lexicons~\cite{depechemoodpp}.
The CS method applies a projection of moral values from a document to its words. The underlying assumption is that each word is associated with the moral value present in the documents where the word appears more frequently. 
To generate a word-by-moral association matrix ($ M_{WM}$), we first create a document-by-moral matrix $M_{DM}$, which shows the distribution of the liberty foundation across the training dataset.
We then generate a word-by-document matrix $M_{WD}$, which indicates the number of occurrences for each word in the vocabulary within a given document, normalized by the total number of words per document.
To obtain the word-by-moral matrix, we perform a multiplication using the following expression:
\begin{equation}
M_{WM} = M_{WD} \cdot M_{DM}
\end{equation}
 
\noindent Using this approach, words and their corresponding value of liberty can be merged by calculating the product of the weight of a word and the weight of the moral value in each document.
The resulting scores are then normalised (column-wise), over-representation issues are addressed, and each lemma is scaled (row-wise) to sum up to one.
Previous validation of lexicons has shown that this normalisation approach is suitable~\cite{depechemoodpp,araque2022libertymfd}.
This approach is referred to as the \textit{CS model} and we generate one lexicon for each dataset except Vaccine.

\paragraph{Liberty Moral Lexicon (LML).}
\label{par:overlap}
The domain-specific lexicons capture topic-dependent expressions of the liberty dimension based on their source datasets. However, we are interested in deriving a general, higher-level representation, so that the final users of the resource have a unified and domain-independent resource.
For this purpose, we synthesise a unified resource merging the obtained discrete lexicons, obtaining our \textit{Liberty Moral Lexicon}. This approach (i) augments the coverage of the consolidated lexicon and (ii) discards uncommon tokens and their annotations.
The basic process to obtain such a lexicon starts by defining a unified vocabulary as the union of the vocabularies of all individual lexicons. This union can be controlled with a selection parameter expressed as a percentage value.
That is, if we define a selection parameter of $m$\%, a word would be included in the union of vocabularies if it appears in at least the $m$\% of all considered lexicons.
Then, we align the numeric assignment each token has in the individual lexicon.
We estimate the average score of these assignments, incorporating them into the unified LML resource\footnote{The Liberty Moral Lexicon (LML) will be publically released upon acceptance.}, if the volume of annotations satisfies the threshold stipulated by the chosen proportion (selection parameter).

\subsection{Evaluation}
\label{subsec:evaluation-design}
To evaluate the performance of the generated resources we designed a wide array of supervised classification tasks.

\paragraph{In-domain evaluation.} We analyze the in-domain performance of our lexicons by testing them on a left-out set of the datasets they are generated from (consisting of 20\% of the original data). Notice how, depending on the dataset, the task is slightly different due to the different type of annotations:
(i) for the BLM and Elect. datasets, the classifier should predict whether the document expresses notions of liberty/oppression or is neutral;
(ii) for the Reddit and WikiCon datasets, it should predict whether a document expresses the libertarian or the conservative point of view..
%
To avoid overfitting, each model is training on the training set of the respective balanced dataset (see Table~\ref{tab:datasets}), leaving the test set for evaluation.

\paragraph{Out-of-domain evaluation.} We perform a series of out-of-domain experiments, testing how well the lexicons can generalize to different domains. In particular, we measure the performance of:
(i) the lexicons generated from Reddit and WikiCon used on the BLM and Election datasets;
(ii) the lexicons generated from BLM and Elect. used on the Reddit and WikiCon datasets;
(iii) the lexicons generated from BLM and Elect., trained on BLM/Elect. and tested on the Vaccine dataset.

Here the different annotation schemes and domains could potentially pose a bigger challenge for the classifier; however, while in the first two cases we can expect to see the impact of the different vocabulary, the train/test split is still coming from the same dataset (in other words, the model has to learn a task using sub-optimal features, but having ``coherent'' data for training and testing).
The Vaccine dataset is instead used to test the out-of-domain performance of the lexicons when the annotation scheme of the evaluation dataset (i.e., the presence or absence of liberty foundation) is coherent with the annotation scheme of the dataset from which the lexicons are generated (BLM and Elect.), but no ideal training data is available (as we are training on an annotated dataset -BLM or Elect- that is different from the Vaccine test dataset). 
%
%
For all experiments, we used logistic regression~\cite{alpaydin2020introduction} and represented each document as a vector of the same length as each lexicon vocabulary, as described in the following.  
Each document is represented by a vector of equal size to the lexicon. For those tokens in the document present in the lexicon, the vector contains the respective polarity score, otherwise zero.
Since this type of representation dramatically simplifies the linguistic information present in the document, we enhance the classification design with two more experiments. We extend each vector representation with the ``statistical summary'' functions, namely the average, maximum, median, variance, and a peak-to-peak score of the lexicon values of that document. This offers the learning models a more complete view of the text.
\paragraph{Combined lexicons.} To test whether it is possible to obtain a more ``general purpose model'', we evaluate two ways of combining the information coming from the different lexicons:
(i) the \textit{LML} described in Section \ref{par:overlap}, which averages the values for words that appear in multiple lexicons;
(ii) the \textit{combined representation}, which is \textit{not} a lexicon, but a method of learning a unified representation by taking into account all available lexicons. 
The advantage of the first method is its simplicity, and that it results in an interpretable lexicon.
On the other hand, the combined representation allows a learning model to observe simultaneously all information contained in the individual lexicons (including words not shared among them); the model might then be able to exploit existing interactions among them.

While the strength of this second approach is that it provides a comprehensive representation obtained through all individual lexicons, overfitting may occur given the large dimensionality of representation. To avoid such issue, we include a feature selection mechanism in the learning model so that the dimension of the feature vector can be reduced. Our approach is based on the Singular Value Decomposition technique (SVD)~\cite{halko2011structure} for transforming the representation of a single lexicon into a continuous vector, which is then input to a machine learning algorithm, in our case a logistic regressor. 
Since these two approaches take into account all generated lexicons, we can consider the results of the LML and combined representation methods:
(i) an in-domain evaluation when applied to the BLM, Elect., Reddit or WikiCon datasets (since they are used to generate the individual lexicons), 
(ii) an out-of-domain evaluation, when applied to the Vaccine dataset (since this is only used as a test set).

\paragraph{Baselines.} As baseline models, we train two classifiers using a unigram representation that includes a frequency-selected vocabulary of sizes 1,000 and 10,000 tokens respectively, which are comparable to the size of the lexicons (Table \ref{tab:lexicon-vocabs}). 
We benchmark LML against the only publicly-available liberty lexicon, namely the LibertyMFD \cite{araque2022libertymfd} as an external baseline.

\paragraph{Lexicon ranking.} To assess the general quality of the lexicons and obtain a ranking of their performance, we performed the Friedman statistical test over all the evaluation results~\cite{demvsar2006statistical}.
In the Friedman test a lower ranking implies a better result for a certain method in comparison to the rest.
In case of ties, these are resolved by averaging the obtained ranks.
The Friedman test has been performed with $\alpha = 0.05$, rejecting the null hypothesis.
We report the macro-averaged F-score as well as the Friedman rank for the evaluation of each resource.



\begin{table}[!b]
 \centering
 \addtolength{\tabcolsep}{1em} 
 \begin{tabular}{lrr}
 \toprule
 \textbf{Lexicon source} & \textbf{Tok. count CS} & \textbf{Tok. count WE}\\
 \midrule
 BLM & 724 & 6,764 \\
 Elect. & 1,994 & 8,777 \\
 Reddit & 10,881 & 63,965 \\
 WikiCon & 61,859 & 62,564 \\
LML & \multicolumn{2}{c}{22,391} \\
 \bottomrule
 \end{tabular}
 \addtolength{\tabcolsep}{2pt} 
 \caption{Number of tokens per lexicon generated for each method.}
 \label{tab:lexicon-vocabs}
\end{table}

\section{Results \& Discussion}

\subsection{Lexicon Generation}

Table~\ref{tab:lexicon-vocabs} shows the vocabulary size for all the generated lexicons.
As previously mentioned, we generated one lexicon from each of the datasets (see Table~\ref{tab:datasets}) using the two proposed methods WE and CS, except for the Vaccine dataset, which is used only for the out of domain evaluation. 
Finally, using the overlapping approach, we obtained a representation that combines the shared tokens from the individual lexicons, using their average scores related to liberty.
For the lexicons generated with the Compositional Semantics method, we applied a 10 frequency cut-off for Reddit and WikiCon.
Due to the limited number of annotated instances in the BLM and Election datasets, we have set a 6 and 3 frequency cut-off, respectively. 
These cut-off variations have been experimentally validated on the training data, and are in line with the literature \cite{depechemoodpp}.
For the lexicons generated with the WE method, the same frequency cut-off has been applied to the Reddit and WikiCon lexicons, while we did not apply any cut-off for the BLM and Election lexicons generated this way, to increase their vocabulary size.
Generally, we have observed that the two methods show a dependency between the number of annotated instances in the training data and the resulting vocabulary size.

For the overlapping lexicons and due to space limitations, we report the data for the lexicon generated using as selection (cut-off) parameter 40\%, which resulted in the best combination in the supervised evaluation.
To do this, we evaluated the selection parameter in the range [10\%, 20\%, ..., 100\%] on the train sets of the considered datasets using 10-fold cross-validation.
This selection justifies the more limited number of tokens with respect to the aggregation of all the lexicons' tokens.

\subsection{Lexicon Evaluation}


\begin{table*}[!b]
 \centering
 \begin{tabular}{lcccc|cc|c}
 \toprule
 & \textbf{BLM} & \textbf{Elect.} & \textbf{Reddit} & \textbf{WikiCon} & \textbf{Vaccine} & \textbf{Vaccine} & \textbf{Friedman} \\ 
 &&&&&(BLM)&(Elect.)&\textbf{Rank}\\
 \textbf{Features:} &&&&&&&\\
 
 \midrule
 Unigram (1000) & 50.90 & 50.12 & 66.80 & 83.84 & 52.11 & 43.66 & 8.1 \\ 
 Unigram (10,000) & 51.84 & 51.81 & 68.39 & 88.10 & 51.17 & 49.70 & 7.6 \\ 
 \midrule
 LibertyMFD (CS) & 50.01 & 50.74 & 64.90 & 88.39 & 42.32 & 47.46 & 9.6 \\
 LibertyMFD (WE) & 50.39 & 51.62 & 65.96 & 88.47 & 39.66 & 48.63 & 8.0 \\
 \midrule
 BLM (CS) & 51.53 & 48.38 & 61.99 & 81.43 & 47.12 & 33.33 & 10.0 \\ 
 Elect. (CS) & 51.21 & 50.52 & 63.59 & 84.89 & 41.44 & 35.67 & 12.6 \\ 
 Reddit (CS) & 52.20 & 49.78 & \textbf{69.01} & 89.96 & 48.82 & 52.39 & 3.8 \\ 
 WikiCon (CS) & 50.24 & 49.65 & 68.12 & \textbf{90.34} & 52.63 & 42.10 & 6.6 \\ 
 \midrule
 BLM (WE) & 46.20 & 53.23 & 65.37 & 85.56 & 52.06 & 57.78 & 9.0 \\ 
 Elect. (WE) & 49.76 & 47.50 & 66.84 & 88.68 & 49.15 & \textbf{58.15} & 6.0 \\ 
 Reddit (WE) & 51.19 & \textbf{55.32} & 64.32 & 88.50 & 53.72 & 54.85 & 8.8 \\ 
 WikiCon (WE) & 50.67 & 53.58 & 65.84 & 88.64 & 51.86 & 54.09 & 7.4 \\ 
 \midrule
 LML & 52.40 & 50.63 & 67.54 & 89.29 & 52.44 & 42.91 & 4.2 \\ 
 \midrule
 Combined Repr. & \textbf{54.14} & 53.84 & 68.12 & 90.26 & \textbf{54.42} & 53.81 & \textbf{3.0} \\
 \bottomrule
 \end{tabular}
 \caption{Unified F1-macro scores and Friedman ranks. Each model is trained on feature sets estimated by the lexicon reported on each row, with training and testing done on the datasets reported in the column name. The ``Vaccine (BLM)'' and ``Vaccine (Elect.)'' columns are the results training on BLM/Elect. and testing on the balanced vaccine dataset, while the features are extracted from the lexicons of each row. Friedman rank shows the best to the worse performing model overall experiments (lower is better). In bold we indicate the lexicon that provides the most discriminatory features for each scenario.}
 \label{tab:unified-supervised}
\end{table*}

Table~\ref{tab:unified-supervised} reports the results of the evaluation.
As described (see Sect.~\ref{subsec:evaluation-design}), for each dataset, we extract linguistic features employing each of the generated lexicons and employ those to train a logistic regression model per dataset. Then we employ the obtained model to infer the liberty moral class of the respective test set.

\paragraph{In-domain evaluation.} Our expectation would be that the models trained on features emerged from lexicons generated on the respective data source would outperform the rest. However, we notice that this is not true in most cases, except when models trained with the Reddit and WikiCon train sets on features extracted from the Reddit and WikiCon lexicons respectively are employed to distinguish between notions of liberty or oppression. 

Looking at these results, it can be seen that generally the learning models trained on the CS lexicon features improve over the unigram baselines, showing that these lexicons capture useful representations. In particular, CS lexicons are consistently on-par or above the baseline, when the lexicon and the dataset are coherent (e.g. lexicon features generated from the BLM dataset, trained and tested to predict the BLM dataset annotations).

In contrast, the overlap lexicon approach, combining information from different lexicons, shows a fairly consistent performance across all datasets.
These results indicate that the difficulty of the task calls for combined knowledge, since enriching the representations with linguistic information from different contexts and writing styles improves the recognition of the liberty moral value in text.

We obtain further confirmation for this hypothesis by looking at the Friedman test: when considering all evaluation combinations, the Combined Representations ranks as the best approach, followed by the LML.
This is to be expected, as this method makes use of \textit{all} lexicons simultaneously, and learns internal representations that can be exploited by a machine learning model.

\paragraph{Out-of-domain evaluation.} 
At the level of individual lexicons, the Reddit one generated with the Compositional Semantics method achieves very good performance overall (being the third best performing lexicon in the Friedman rank) also when used on non-Reddit datasets. This may be an effect of the larger number of tokens and general quality of the original dataset, which probably includes a richer vocabulary for a variety of topics discussed by libertarians and conservatives.
Besides, this observation offers the insight that, even though the annotations of the used lexicon are not completely aligned (e.g., using the Reddit lexicon for predicting Liberty/Oppression, while the Reddit dataset from which it is generated captures the Libertarian/Conservative divide), the knowledge captured by the lexicon can aid in the classification task. A possible explanation is that the lexicons cover the whole gamut of association strengths, thus capturing a balanced view of language and not just words strongly correlated with the \textit{liberty} foundation; this could help the classifier learn the threshold between documents expressing this foundation (which will have more words with ``extreme'' values) and those which do not (probably consisting of mostly ``neutral'' words).

Regarding the ``stricter'' out-of-domain evaluation, the right side of Table~\ref{tab:unified-supervised} reports the results obtained when models fed with features extracted by the BLM, Elect, WikiCon, and Reddit lexicons (either CS or WE), are trained on the BLM and Elect. training dataset and tested on the Vaccine dataset. 
We notice that, again, the models trained on the Overlapping Lexicon are consistently performing well, while feature extraction based on individual lexicons led to models that did not consistently outperform the baseline.

This experiment offers interesting insights into the generalization capabilities of the proposed method.
Although, the absolute best performance is obtained with the Elect. (WE) lexicon trained on the Election dataset, is not a generalisable finding; the same lexicon trained on the BLM data fails to outperform the baseline. 
Overall, this design offers insights on the adaptability and generalization capability of the lexicons.
Interestingly, the combined representation approach ranks always high validating the fact that combining knowledge from different base lexicons does improve the understanding of the liberty foundation.

\paragraph{Domain Specific Insights.}

A recurrent pattern is that the overlapping lexicon outperforms the other lexicons in both in-domain and out-of-domain experiments. 
To gain more insights on the effect of the social context on the moral nuances a specific lemma may have, we employed the TOMEA approach proposed in \cite{liscio2023}.
According to TOMEA, the overlapping lexicon differs the most with respect to the BLM lexicons generated by both the CS and the WE methods with scores .17 and .13, respectively. 
Glancing into the most distant words, we have 
``fake'',
``lawmaker'',
``elected'',
``supporters'',
``antifa'',
``openly'',
``tweets'',
``sympathizer'',
``tyranny'',
``globalist'',
``dead'', to be considered more oppressive than the average lexicons in the individual BLM lexicons than in the others.
Such domain specific nuances may be important when analysing a specific argument but can introduce biased in the models when analysing broader subjects. TOMEA to this respect is a valuable method to gain insights and foster the transparency and accountability of the findings.

\section{Conclusion}

The aim of this study was to provide a lexical resource for the moral foundation of liberty able to generalise across various domains. The ``Liberty/Oppression'' foundation expresses people's inclinations towards autonomy and their resistance to dominion. Our contributions to the current state of the art are manifold. 
Firstly, we provide an enriched version of the MFTC annotations for the BLM and ELECT datasets as per the liberty foundation. 
Further, we generated a series of lexicons with two complementary approaches and thoroughly evaluated them via a series of both in- and out-of domain experimental scenarios. 
Aside the individual lexicons, we also proposed a final version combining the information from both approaches which is the resource we propose as the final \textit{Liberty Moral Lexicon (LML)}. 
This combined representation exploits information in all generated lexicons, thus offering a classifier a more comprehensive representation outperforming individual expert lexicons \cite{hill2025wisdom}.
This resource showed solid generalisability potentials.
As seen in our experimental evaluation, the generated resources capture relevant knowledge that can be leveraged to assessing liberty in texts with statistical significance.
Finally, by employing the TOMEA method, we provide insights into the dynamics of linguistic variability according to the context in which the word is used.
This study reaffirms the critical role of lexicons in advancing moral foundation analysis by enabling context-aware evaluations of concepts like liberty. Tools like TOMEA rely on these structured resources to decode the domain-specific expressions of moral values effectively and are fundamental in fields related to communication or policymaking when accountability and transparency are a strong requirement. 


\section*{Acknowledgments}
We would like to thank the colleagues of the 
Monitoring, Indicators \& Impact Evaluation Unit of the European Commission Joint Research Centre 
for their support. The views expressed are purely those of the authors and may not in any circumstance be regarded as stating an official position of the European Commission. 
OA would like to acknowledge the funding of
funded by the Spanish Ministry for Economic Affairs and Digital Transformation and by the European Union - NextGenerationEU within the Programme UNICO I+D Cloud (TSI-063100-2022-0002);
as well as the funding of the project CPP2023‐010437 financed by the MCIN / AEI /
10.13039/501100011033 / FEDER, UE.
This work has been supported by the Madrid Government (Comunidad de Madrid-Spain) under the Multiannual Agreement 2023-2026 with Universidad Politécnica de Madrid in the Line A, Emerging PhD researchers (project MORA, DOCTORES-EMERGENTES-24-9UMLXZ-37-IIGW).
KK acknowledges support from the Lagrange Project of ISI Foundation funded by CRT Foundation.

%


\begin{footnotesize}
\bibliographystyle{splncs04}
\bibliography{LOD/refs.bib}
\end{footnotesize}


%
%
%
%
\end{document}